\documentclass[11pt]{article}

\usepackage[preprint]{acl}

\usepackage{times}
\usepackage{latexsym}
\usepackage[T1]{fontenc}
\usepackage[utf8]{inputenc}
\usepackage{microtype}
\usepackage{inconsolata}
\usepackage{graphicx}
\usepackage{booktabs}
\usepackage{amsmath}
\usepackage{amssymb}
\usepackage{multirow}
\usepackage{xcolor}
\usepackage{enumitem}
\usepackage[most]{tcolorbox}

\newcommand{\fwt}{\mathrm{FWT}}
\newcommand{\bwt}{\mathrm{BWT}}
\newcommand{\drr}{\Delta\mathrm{RR}}
\newcommand{\dft}{\Delta\mathrm{NL}}

\newtcolorbox{casebox}[1]{
  enhanced,
  breakable,
  colback=white,
  colframe=black!20,
  boxrule=0.4pt,
  arc=1mm,
  left=2mm,
  right=2mm,
  top=1.5mm,
  bottom=1.5mm,
  title={#1},
  fonttitle=\bfseries,
  coltitle=black
}

\title{When Continual Learning Moves to Memory:\\A Study of Experience Reuse in LLM Agents}

\author{Qisheng Hu \quad Quanyu Long \quad Wenya Wang \\
        Nanyang Technological University \\
        \texttt{qisheng001@e.ntu.edu.sg}
        }

\begin{document}
\maketitle

\begin{abstract}
Memory-augmented LLM agents offer an appealing shortcut to continual learning: rather than updating model parameters, they accumulate experience in external memory, seemingly sidestepping the stability–plasticity dilemma of parametric learning. We show that this challenge does not disappear but resurfaces at the memory level. Under a limited context window, old and new experiences compete during retrieval, relocating the continual-learning bottleneck from parameter updates to memory access. To study this phenomenon, we introduce a $(k,v)$ framework that disentangles two fundamental design axes of external memory: how experience is \emph{represented} and how it is \emph{organized} for retrieval. Across sequential-task experiments in ALFWorld and BabyAI, we find that abstract procedural memories transfer more reliably than detailed trajectories, while negative transfer disproportionately harms the hard cases. Moreover, finer-grained memory organization is not universally beneficial: designs that yield strong forward transfer can simultaneously induce severe forgetting. Together, these results reveal that external memory does not resolve the continual-learning problem; it reshapes it into a problem of memory representation and retrieval design.
\end{abstract}

\section{Introduction}
\label{sec:intro}

A growing body of work equips LLM agents with external memory, from retrieval-augmented language models \citep{borgeaud2022retro,wang2023longmem} to agent systems that store and reuse experience across interactions \citep{zhong2024memorybank,shinn2023reflexion,wang2023voyager,packer2023memgpt,zhao2024expel,xu2025amem}. Such designs are appealing because they appear to sidestep a central challenge of learning from sequential experience: rather than forcing old and new knowledge into the same model weights, the agent can preserve and reuse past experience in an external memory and continue accumulating new experience online.
This makes memory-augmented agents especially relevant to continual learning (CL), whose core setting assumes sequentially arriving data and ongoing adaptation over time. In this sense, the online nature of LLM agent memory is not merely compatible with CL but inherently instantiates its streaming requirement.
Moreover, because old experience can remain stored while new experience is continuously added, one might expect memory-augmented agents to enjoy both strong \emph{stability}, in the sense of retaining previously acquired capability through stored memories, and strong \emph{plasticity}, in the sense of adapting to new tasks by incorporating new experience into memory.

\begin{figure}[t]
  \centering
  \setlength{\belowcaptionskip}{-0.5cm}
  \includegraphics[width=1.0\columnwidth]{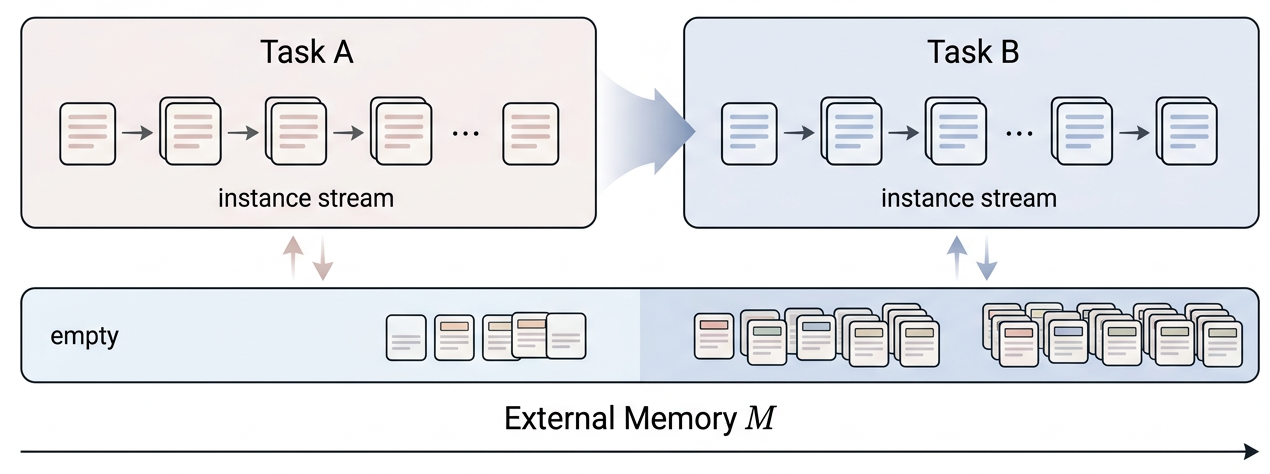}
  \caption{Evaluation protocol for the $A\to B$ direction. The agent first learns a stream of Task~A instances from empty memory, then continues with a stream of Task~B instances while reusing the accumulated memory.}
  \label{fig:cl-paradigm}
\end{figure}

However, external memory does not eliminate the continual-learning challenge but instead relocates it.
In parametric CL, the central difficulty is interference in weight space: learning new tasks can overwrite knowledge useful for old ones, leading to the well-known stability--plasticity dilemma \citep{kirkpatrick2017ewc,lopez2017gradient,wang2024clsurvey}.
In memory-augmented agents, past experience can in principle be \emph{stored} without overwriting parameters, but old experience is only useful if it can be effectively \emph{retrieved} and inserted into the model's limited context window at the right time.
Once tasks arrive sequentially, old and new experiences therefore still compete: the continual-learning bottleneck shifts from how to store knowledge to which experience is retrieved, reused, and prioritized under a limited retrieval budget.
Concretely, irrelevant memories may be recalled, introducing \emph{retrieval pollution}; useful experiences may be displaced by other retrieved items under a finite context window, creating \emph{context competition}; and as the memory store grows across tasks, relevant experience may become harder to identify, resulting in \emph{memory dilution}.

Despite rapid progress in memory design, we still lack a controlled understanding of how external memory behaves under sequential task shift.
Recent surveys organize the design space of memory-equipped agents \citep{zhang2025memorysurvey}, and recent work has begun to frame richer retrieval as a form of non-parametric continual learning \citep{gutierrez2025ragmemory}.
However, much of the current literature primarily demonstrates improved performance within a task or a task family, rather than evaluating the core questions that define continual learning as a learning problem: does accumulated experience support transfer to future tasks, or does it induce negative transfer? Does new experience preserve or degrade performance on earlier tasks?
In other words, although many memory-agent systems are continuous in operation, far fewer study continual learning in the classical sense.

In this paper, we study memory-augmented online learning through a continual-learning lens.
Using controlled task sequences in ALFWorld \citep{shridhar2021alfworld} and BabyAI \citep{chevalier2019babyai}, we measure not only within-task improvement but also transfer and forgetting across tasks, focusing on two fundamental design axes: how experience is \emph{represented} (from raw episodic trajectories to abstract procedural insights) and how it is \emph{organized} (memory granularity and retrieval frequency).
We further leverage subset-level diagnostic analyses that localize where negative transfer arises. Our central findings are:
\begin{itemize}[nosep,leftmargin=*]
\item \textbf{Abstraction shapes whether memory helps or hurts.} Raw trajectories tend to hinder adaptation to new tasks, while abstract insights are safer to reuse and can reduce forgetting of earlier tasks.
\item \textbf{Negative transfer concentrates on hard cases.} Cross-task retrieval disproportionately harms cases the agent cannot yet solve on its own, while cases it already handles are comparatively robust.
\item \textbf{Finer-grained memory organization is not universally better.} Both finer storage granularity and more frequent retrieval can help or hurt, depending on the diversity of stored experience and the structure of the target task.
\item \textbf{A stability--plasticity trade-off can emerge through retrieval.} The design that well supports adaptation to a new task can simultaneously cause significant forgetting of an earlier one.
\end{itemize}

\section{Framework \& Experimental Setup}
\label{sec:framework}

\subsection{Agent Memory as Continual Learning}

Our experimental paradigm consists of two sequential phases, Task~A and Task~B, each containing a stream of task instances.
Figure~\ref{fig:cl-paradigm} illustrates the process: the agent starts from empty memory(experience pool), processes Task~A instances, and then continues with Task~B while carrying over the accumulated memory.
In our main experiments, each phase contains $n=200$ training instances.
For each instance, the agent retrieves relevant experience from a shared external memory $\mathcal{M}$, attempts the task, and then stores the resulting experience back into $\mathcal{M}$.
When the phase switches from Task~A to Task~B, $\mathcal{M}$ is reused and continues to grow.
Concretely, the agent follows a ReAct-style interaction loop \citep{yao2023react} and uses the ReMe memory module \citep{cao2025remember}; retrieved experiences are inserted into the model context before the first action step.

This setup is naturally a form of \textbf{non-parametric continual learning}.
The LLM backbone is frozen, so there is no parametric forgetting.
However, behavior still depends on what is recalled into the finite context window, so retrieval pollution, context competition, and memory dilution remain central sources of interference.

\begin{table}[t]
  \centering
  \small
  \setlength{\belowcaptionskip}{-0.5cm}
  \setlength{\tabcolsep}{4pt}
  \begin{tabular}{lll}
  \toprule
  & \textbf{Parametric CL} & \textbf{Memory CL} \\
  \midrule
  Carrier     & Model weights      & External memory \\
  Interference & Gradient overwrite & Retrieval pollution \\
  Bottleneck  & Parameter capacity & Context window \\
  \bottomrule
  \end{tabular}
  \caption{Parametric vs.\ memory-based CL. External memory avoids weight overwriting but introduces a retrieval bottleneck.}
  \label{tab:cl-compare}
\end{table}

\subsection{A $(k,v)$ View of Experience Reuse}

Given continual learning in memory-augmented agents hinges on retrieval, the central question is whether stored experience can be effectively \emph{reused} across tasks and over time.
In practice, experience reuse varies widely across memory designs: some memories remain overly specific to transfer well, while others fail to be retrieved when needed.
This suggests that successful reuse depends on two factors: how experience is \emph{represented}, and how it is \emph{organized and accessed}.
To analyze these factors systematically, we model each memory unit as a key--value pair $(k,v)$. This abstraction separates the representation of experience from the mechanism used to retrieve it, allowing us to characterize reuse along two complementary dimensions.

\textbf{Value $v$} captures the \emph{representation} of experience.
A memory may preserve raw, instance-specific procedural detail or store more abstract, transferable insights, which in turn shapes how brittle or reusable the retrieved content is (Study~1).

\textbf{Key $k$} captures how experience is organized and accessed.
We distinguish \emph{storage granularity} (does one memory unit correspond to a whole task-level bundle or a insight-level unit?) from \emph{retrieval granularity} (is memory retrieved once per task instance or updated more frequently during execution?). Together these determine whether the right unit can be matched and recalled at the right time (Study~2).

\subsection{Continual Learning Metrics}

For a two-phase sequence, we use transfer-oriented metrics from continual learning on held-out test sets.
Task~A and Task~B denote the two task families in the sequence.
Here $\text{Acc}_T(\cdot)$ denotes success rate on the held-out test set of task $T$.

\paragraph{Forward Transfer (FWT).}
For the sequence $A\!\to\!B$, forward transfer measures whether prior learning on Task~A helps or hurts Task~B:
\begingroup\small
\[
\fwt(A\!\to\!B) = \text{Acc}_B(\varnothing\!\to\!A\!\to\!B) - \text{Acc}_B(\varnothing\!\to\!B).
\]
\endgroup
$\fwt<0$ indicates impaired plasticity, while $\fwt>0$ indicates positive transfer.

\paragraph{Backward Transfer (BWT).}
For the same sequence $A\!\to\!B$, backward transfer measures whether the later phase degrades the earlier task:
\begingroup\small
\[
\bwt(A\!\to\!B) = \text{Acc}_A(\varnothing\!\to\!A\!\to\!B) - \text{Acc}_A(\varnothing\!\to\!A),
\]
\endgroup
$\bwt<0$ indicates forgetting, while $\bwt>0$ indicates positive backward transfer.
Both FWT and BWT are indexed by the \emph{training sequence} ($A\!\to\!B$ or $B\!\to\!A$): FWT refers to the later task, BWT to the earlier one.

\paragraph{RR/NL subset diagnostic.}
Average success alone does not tell us \emph{where} transfer occurs, so we partition each test set using the no-memory baseline into a baseline-success subset (easy cases) and a baseline-fail subset (hard cases).
We then measure \textbf{Retention Rate (RR)}, the success rate on the baseline-success subset after memory is introduced, and \textbf{New Learning rate (NL)}, the success rate on the baseline-fail subset after memory is introduced.
For a cross-task run $X\to T$, we define:
\begin{equation*}
\begin{aligned}
\drr &= \text{RR}(\varnothing\!\to\!X\!\to\!T) - \text{RR}(\varnothing\to T), \\
\dft &= \text{NL}(\varnothing\!\to\!X\!\to\!T) - \text{NL}(\varnothing\to T).
\end{aligned}
\end{equation*}

These deltas distinguish two failure modes:
\begin{itemize}[nosep,leftmargin=*]
\item $\drr < 0$: \textbf{harmful reuse}. Cross-task memory breaks previously solved cases.
\item $\dft < 0$: \textbf{ineffective reuse}. Cross-task memory fails to help, or further harms, previously difficult cases.
\end{itemize}

\subsection{Environments, Task Pairs, and Protocol}

\begin{table}[t]
\centering
\small
\setlength{\belowcaptionskip}{-0.5cm}
\setlength{\tabcolsep}{3pt}
\begin{tabular}{llccc}
\toprule
\textbf{Env} & \textbf{Task} & \textbf{BL} & $n_{\text{s}}$ & $n_{\text{f}}$ \\
\midrule
\multirow{2}{*}{ALFWorld}
& A: \texttt{pick\_and\_place} & 95\% & 95 & 5\rlap{$^{\dagger}$} \\
& B: \texttt{pick\_clean\_place} & 54\% & 54 & 46 \\
\midrule
\multirow{2}{*}{BabyAI}
& A: \texttt{action\_obj\_door} & 70\% & 70 & 30 \\
& B: \texttt{find\_obj} & 49\% & 49 & 51 \\
\bottomrule
\end{tabular}
\caption{Task pairs. BL: no-memory baseline. $n_{\text{s}}$/$n_{\text{f}}$: baseline-success/fail subset sizes. $^{\dagger}$Small baseline-fail subset ($n<10$), NL-based estimates are noisy and omitted from the main results tables.}
\label{tab:tasks}
\end{table}

We analyze one task pair in ALFWorld \citep{shridhar2021alfworld} and one in BabyAI \citep{chevalier2019babyai} (Table~\ref{tab:tasks}).
In ALFWorld, Task~A uses instructions of the form ``put X in Y,'' while Task~B uses ``put a clean X in Y.'' The two tasks are therefore highly similar at the instruction level but differ procedurally, making them well-suited for studying retrieval pollution.
In BabyAI, Task~A produces heterogeneous queries across three sub-task types, whereas Task~B produces homogeneous ``find [object]'' queries. This contrast is especially useful in Study~2 for analyzing how storage granularity and retrieval frequency interact with query structure.
Further protocol details are given in Appendix~\ref{app:details}.

\section{Study 1: Memory Representation}
\label{sec:study1}

We investigate the value $v$: how does the representation of stored experience affect CL dynamics?
We compare two representations:
\textbf{Raw}, which stores the action--observation trajectory from each episode,
and \textbf{Insight}, which stores LLM-abstracted strategy insights distilled from the trajectory.

\begin{table}[t]
\centering
\small
\setlength{\belowcaptionskip}{-0.5cm}
\setlength{\tabcolsep}{3.2pt}
\begin{tabular}{ll rrr rr}
\toprule
& & \multicolumn{3}{c}{\textbf{Performance (\%)}} & \multicolumn{2}{c}{\textbf{Diagnostic}} \\
\cmidrule(lr){3-5} \cmidrule(lr){6-7}
\textbf{Env} & \textbf{Repr}
  & $\varnothing\!\!\to\!\!T$ & $X\!\!\to\!\!T$ & \textbf{FWT}
  & $\drr$ & $\dft$ \\
\midrule
\multicolumn{7}{l}{\textit{A$\to$B}} \\
\midrule
\multirow{2}{*}{\rotatebox[origin=c]{0}{\scriptsize ALF}}
& Raw      & 80.5 & 71.0 & $-$9.5  & $+$4.6  & $-$26.1 \\
& Insight  & 65.5 & 72.0 & $+$6.5  & $+$9.3  & $+$3.3  \\
\midrule
\multirow{2}{*}{\rotatebox[origin=c]{0}{\scriptsize Baby}}
& Raw      & 53.5 & 46.0 & $-$7.5  & $-$2.0  & $-$12.7 \\
& Insight  & 46.0 & 55.0 & $+$9.0  & $+$8.2  & $+$9.8  \\
\midrule
\multicolumn{7}{l}{\textit{B$\to$A}} \\
\midrule
\multirow{2}{*}{\rotatebox[origin=c]{0}{\scriptsize ALF}}
& Raw      & 92.0 & 91.5 & $-$0.5  & $-$0.7  & \multicolumn{1}{c}{---} \\
& Insight  & 97.0 & 94.5 & $-$2.5  & $-$1.1  & \multicolumn{1}{c}{---} \\
\midrule
\multirow{2}{*}{\rotatebox[origin=c]{0}{\scriptsize Baby}}
& Raw      & 64.0 & 71.0 & $+$7.0  & $+$2.9  & $+$16.7 \\
& Insight  & 65.5 & 73.0 & $+$7.5  & $+$1.4  & $+$21.7 \\
\bottomrule
\end{tabular}
\caption{Study~1 FWT results. ALFWorld B$\to$A: $\dft$ omitted for small samples.}
\label{tab:s1-fwt}
\end{table}

\subsection{Abstraction Facilitates Safer Cross-Task Reuse}

\begin{figure}[t]
\centering
\setlength{\abovecaptionskip}{0.2cm}
\setlength{\belowcaptionskip}{-0.3cm}
\includegraphics[width=\columnwidth]{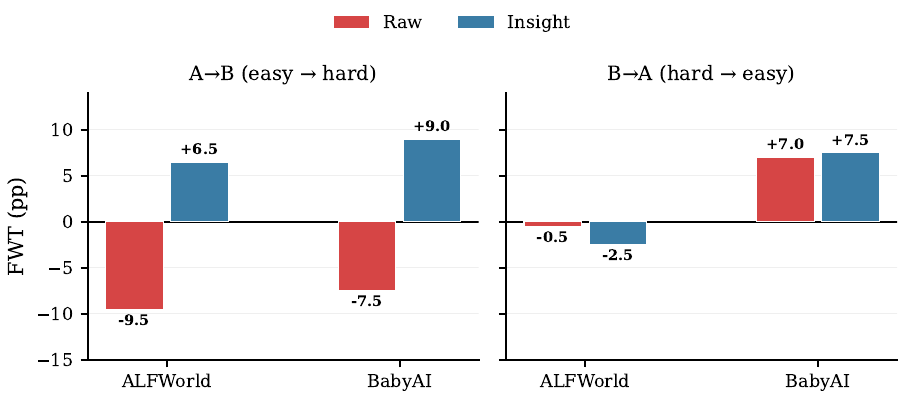}
\caption{FWT by representation. In the A$\to$B direction, Raw produces negative transfer while Insight yields positive transfer across both environments.}
\label{fig:s1-fwt}
\end{figure}

Table~\ref{tab:s1-fwt} and Figure~\ref{fig:s1-fwt} present the full results.
The clearest pattern appears in the A$\to$B sequence: Raw produces negative FWT across both environments, while Insight shifts transfer to positive.
At the task-family level, Task~A is relatively easier than Task~B in both environments, as reflected in the baseline success rates in Table~\ref{tab:tasks}.

This directional asymmetry helps explain why abstraction matters most here.
When experience from a relatively easier task is reused for a relatively harder one, raw trajectories can carry over procedures that look relevant but miss the additional structure the harder task requires.
In ALFWorld, for example, Task~A trajectories prescribe ``take$\to$go$\to$place'' while Task~B requires an additional cleaning step.
In this case, raw procedural detail can push the agent toward a behavior that is locally plausible but globally ill-suited to the new task (see Appendix~\ref{app:case} for qualitative examples).

Insight memory reduces this risk by abstracting away from task-specific action scripts and preserving higher-level guidance.
As a result, abstraction appears to facilitate safer reuse precisely in the direction where cross-task procedural mismatch is most damaging.
By contrast, in the B$\to$A sequence the gap between Raw and Insight becomes much smaller, suggesting that experience from the relatively harder task is less harmful when reused for the easier one.
Appendix~\ref{app:s1-direction} provides a fuller discussion of this directional asymmetry.

\subsection{Negative Transfer Is Concentrated on Hard Cases}

\begin{figure}[t]
\centering
\setlength{\abovecaptionskip}{0.2cm}
\setlength{\belowcaptionskip}{-0.4cm}
\includegraphics[width=\columnwidth]{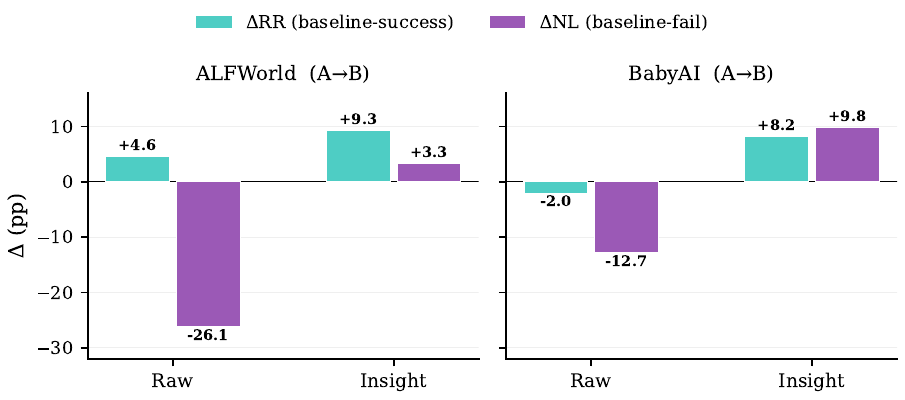}
\caption{$\drr$/$\dft$ decomposition (A$\to$B). Under Raw, the hard subset (purple) suffers more than the easy subset (teal), especially on ALFWorld. Insight memory largely reduces this asymmetry.}
\label{fig:s1-rrft}
\end{figure}

The RR/NL diagnostic (Figure~\ref{fig:s1-rrft}) shows that negative transfer is not evenly distributed.
The most illustrative case is ALFWorld A$\to$B: under Raw, $\drr = +4.6$ but $\dft = -26.1$, so the easy subset is largely preserved while the hard subset degrades sharply.
This pattern is intuitive: cases the agent can already solve tend to remain solvable regardless of what is retrieved, whereas cases it cannot yet solve depend much more on retrieved guidance.
When that guidance comes from a different task, more procedural details is more likely to mislead than to help.

Insight memory largely reduces this vulnerability, as abstract representations do not prescribe task-specific procedures.

In summary, representation choice is especially important for the hardest test instances as they are the most vulnerable to cross-task noise.

\subsection{Within-Task Advantage Does Not Predict Cross-Task Benefit}
\label{sec:s1-reversal}

\begin{figure}[t]
\centering
\setlength{\abovecaptionskip}{0.2cm}
\setlength{\belowcaptionskip}{-0.3cm}
\includegraphics[width=\columnwidth]{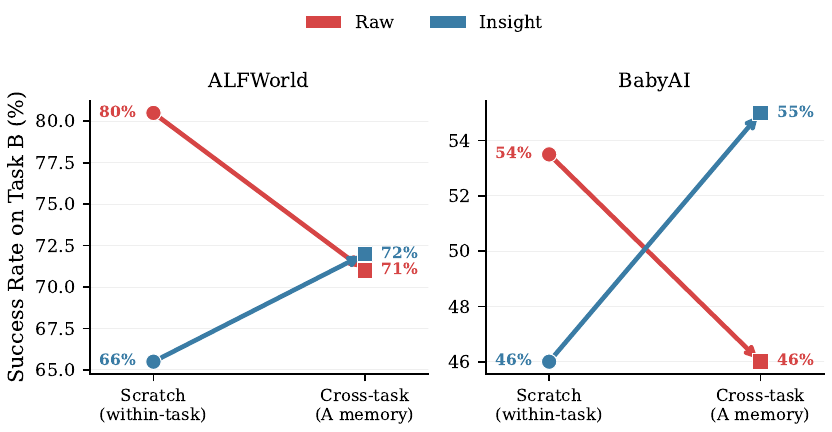}
\caption{Within-task vs.\ cross-task performance on Task~B. Raw's large within-task advantage disappears or reverses in the cross-task setting.}
\label{fig:s1-slope}
\end{figure}

The Raw--Insight comparison highlights a practical pitfall: the representation that works best within a single task distribution may not retain that advantage once cross-task reuse is introduced (Figure~\ref{fig:s1-slope}).
On ALFWorld Task~B, Raw has a large within-task advantage because detailed trajectories are highly informative when reused within the \emph{same} task family.
Under cross-task reuse, however, that advantage disappears: both conditions converge to similar performance, with Insight slightly ahead.
On BabyAI, the shift is stronger still: Raw declines while Insight improves, producing a clear crossover.
The underlying mechanism is similar in both environments: procedural detail is an asset within-task but can become a liability once the retrieved procedure belongs to a different task (see Appendix~\ref{app:case}).
The implication is straightforward: \textbf{a memory design that looks best within a single task distribution may not be the best choice for cross-task deployment}.

\subsection{Backward Transfer: Abstraction Reduces Forgetting}

\begin{figure}[t]
\centering
\setlength{\abovecaptionskip}{0.2cm}
\setlength{\belowcaptionskip}{-0.4cm}
\includegraphics[width=\columnwidth]{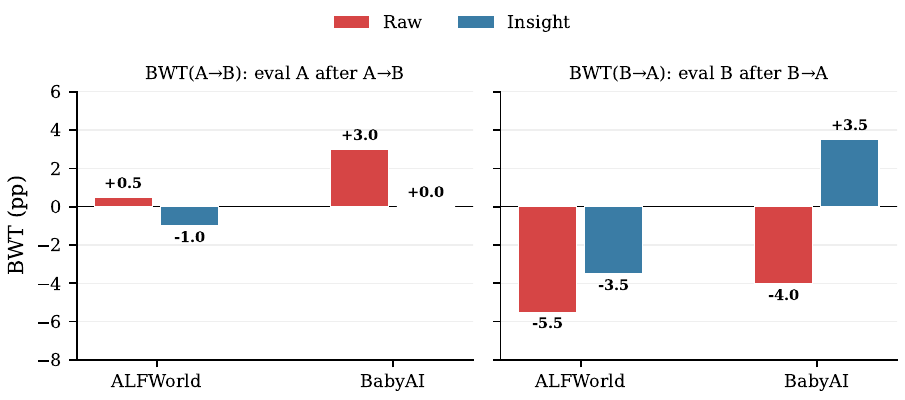}
\caption{BWT by representation. On BabyAI, Insight achieves positive BWT under the B$\to$A sequence while Raw causes forgetting. On ALFWorld, both representations show moderate forgetting.}
\label{fig:s1-bwt}
\end{figure}

\begin{table}[t]
\centering
\small
\setlength{\tabcolsep}{3.5pt}
\begin{tabular}{ll rr rr}
\toprule
& & \multicolumn{2}{c}{\textbf{BWT(A$\to$B)}} & \multicolumn{2}{c}{\textbf{BWT(B$\to$A)}} \\
\cmidrule(lr){3-4} \cmidrule(lr){5-6}
\textbf{Env} & \textbf{Repr} & BWT & $\drr$ & BWT & $\drr$ \\
\midrule
\multirow{2}{*}{ALF}
& Raw      & $+$0.5 & $+$0.5 & $-$5.5 & $+$2.9 \\
& Insight  & $-$1.0 & ~~0.0  & $-$3.5 & $+$4.6 \\
\midrule
\multirow{2}{*}{Baby}
& Raw      & $+$3.0 & $+$2.9 & $-$4.0 & $+$2.1 \\
& Insight  & ~~0.0  & $+$2.2 & $+$3.5 & $+$8.2 \\
\bottomrule
\end{tabular}
\caption{Study~1 backward transfer. BWT($X{\to}Y$) re-evaluates Task~$X$ after the full $X{\to}Y$ sequence. $\drr$ shown for the baseline-success subset.}
\label{tab:s1-bwt}
\end{table}

We now ask whether the same abstraction effect extends to \emph{stability}, measured via backward transfer (Table~\ref{tab:s1-bwt}, Figure~\ref{fig:s1-bwt}).

BWT(A$\to$B) is near-zero in all conditions, which is unsurprising given Task~A's high baseline and limited room for change.
The more informative sequence is B$\to$A, where BWT measures whether continuing with Task~A degrades previously learned Task~B performance.
On ALFWorld, both representations show moderate forgetting.
On BabyAI, however, the two diverge: Insight achieves \emph{positive} backward transfer, while Raw causes forgetting.

This mirrors the forward-transfer story.
Raw trajectories from Task~A are task-specific procedures that, once mixed into the pool, can mislead retrieval for Task~B.
Insight memories from Task~A, by contrast, encode task-agnostic principles that may also be useful for Task~B, enriching rather than polluting the pool.
The benefit of abstraction therefore appears to extend beyond forward transfer: \textbf{it can also reduce forgetting, and in some settings even support positive backward transfer}.

\subsection{Study 1 Summary}

\begin{enumerate}[nosep,leftmargin=*]
\item \textbf{Abstraction facilitates safer cross-task reuse.} Raw trajectories tend to behave like misleading procedures across task boundaries, while abstract insights are more transferable.
\item \textbf{Negative transfer is concentrated where help is most needed.} Cross-task noise disproportionately harms hard cases, which depend most on retrieved guidance.
\item \textbf{Within-task evaluation is not enough.} A representation that looks strong within one task distribution can lose its advantage once cross-task reuse is introduced.
\item \textbf{Abstraction can also reduce forgetting.} Beyond helping adaptation to new tasks, abstract insights make previously learned behavior less vulnerable to cross-task interference.
\end{enumerate}

\section{Study 2: Memory Organization}
\label{sec:study2}

\begin{figure}[t]
  \centering
  \includegraphics[width=0.85\columnwidth]{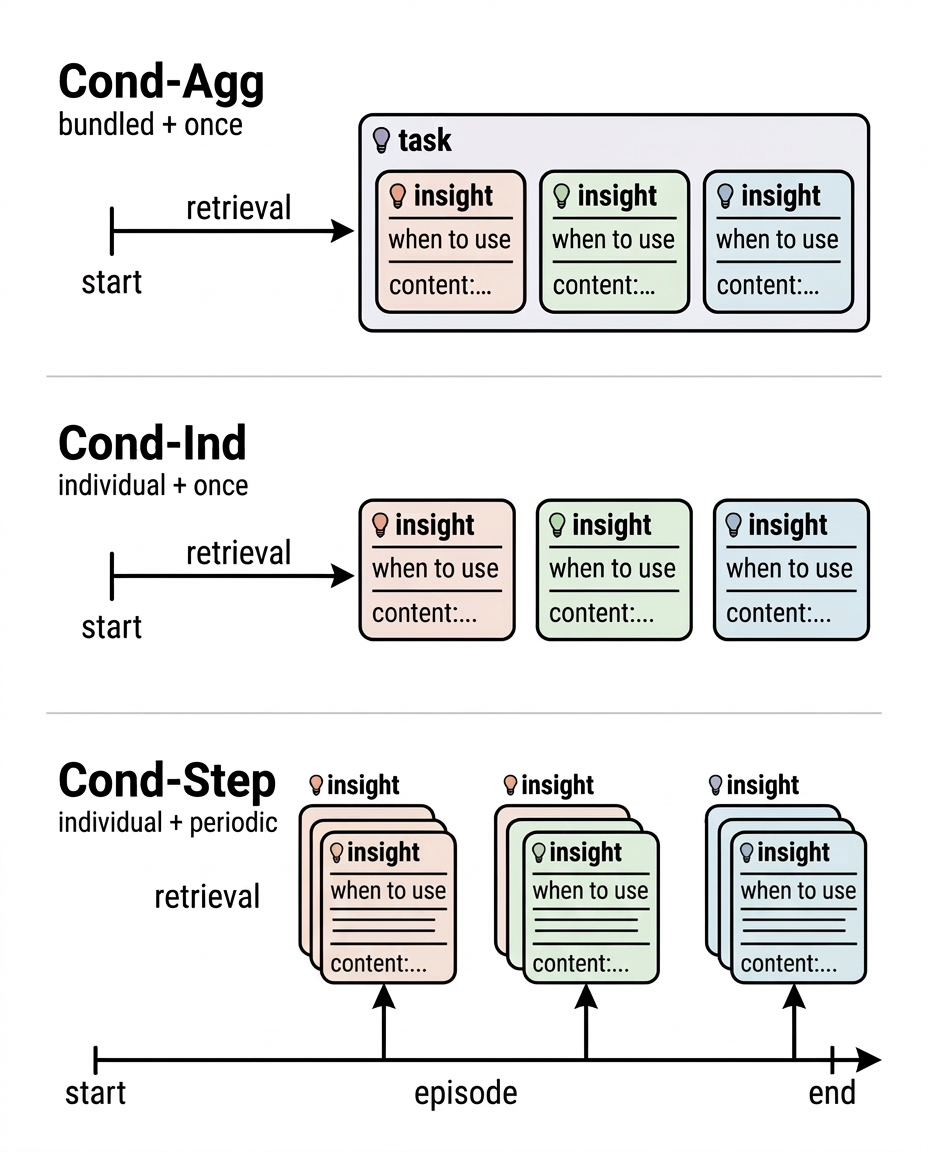}
  \caption{Visual comparison of Cond-Agg, Cond-Ind, and Cond-Step.}
  \label{fig:s2-cond-comparison}
\end{figure}

Study~1 fixed memory organization and varied representation.
We now fix representation to Insight (the better $v$ from Study~1) and vary the $k$ dimension: how memory units are organized and accessed.
We vary two aspects of granularity (Figure~\ref{fig:s2-cond-comparison}):

\textbf{Cond-Agg} bundles all insights from an episode into a single memory entry keyed by the task instruction. The agent queries once at the start of each episode and retrieves the full bundle.
\textbf{Cond-Ind} stores each insight as a separate entry with its own retrieval key (a short applicability description). The agent still queries once per episode but can retrieve individual insights from different source episodes.
\textbf{Cond-Step} uses the same insight-level storage as Cond-Ind but re-queries periodically during execution, allowing retrieved guidance to update as the episode progresses.
Appendix~\ref{app:examples} shows examples of stored entries for each condition.

Comparing Agg vs.\ Ind isolates \emph{unit granularity} (bundled vs.\ individual storage), while Ind vs.\ Step isolates \emph{retrieval frequency} (single vs.\ periodic querying). All conditions surface roughly the same number of insights per retrieval, keeping the information budget comparable.

\begin{table}[t]
\centering
\small
\setlength{\tabcolsep}{3pt}
\begin{tabular}{ll rrr rr}
\toprule
& & \multicolumn{3}{c}{\textbf{Performance (\%)}} & \multicolumn{2}{c}{\textbf{Diagnostic}} \\
\cmidrule(lr){3-5} \cmidrule(lr){6-7}
\textbf{Env} & \textbf{Cond}
  & $\varnothing\!\!\to\!\!T$ & $X\!\!\to\!\!T$ & \textbf{FWT}
  & $\drr$ & $\dft$ \\
\midrule
\multicolumn{7}{l}{\textit{A$\to$B}} \\
\midrule
\multirow{3}{*}{\rotatebox[origin=c]{0}{\scriptsize ALF}}
& Agg  & 65.5 & 72.0 & $+$6.5  & $+$9.3  & $+$3.3  \\
& Ind  & 55.0 & 53.0 & $-$2.0  & $-$4.6  & $+$0.9  \\
& Step & 79.5 & 78.0 & $-$1.5  & $-$3.7  & $+$1.1  \\
\midrule
\multirow{3}{*}{\rotatebox[origin=c]{0}{\scriptsize Baby}}
& Agg  & 46.0 & 55.0 & $+$9.0  & $+$8.2  & $+$9.8  \\
& Ind  & 32.0 & 47.0 & $+$15.0 & $+$19.4 & $+$10.8 \\
& Step & 48.0 & 45.5 & $-$2.5  & $-$4.1  & $-$1.0  \\
\midrule
\multicolumn{7}{l}{\textit{B$\to$A}} \\
\midrule
\multirow{3}{*}{\rotatebox[origin=c]{0}{\scriptsize ALF}}
& Agg  & 97.0 & 94.5 & $-$2.5  & $-$1.1  & \multicolumn{1}{c}{---} \\
& Ind  & 96.5 & 97.0 & $+$0.5  & $+$0.5  & \multicolumn{1}{c}{---} \\
& Step & 99.0 & 98.5 & $-$0.5  & $-$1.1  & \multicolumn{1}{c}{---} \\
\midrule
\multirow{3}{*}{\rotatebox[origin=c]{0}{\scriptsize Baby}}
& Agg  & 65.5 & 73.0 & $+$7.5  & $+$1.4  & $+$21.7 \\
& Ind  & 75.5 & 74.5 & $-$1.0  & ${\approx}$0 & $-$1.7  \\
& Step & 70.0 & 77.5 & $+$7.5  & $+$4.3  & $+$15.0 \\
\bottomrule
\end{tabular}
\caption{Study~2 Forward transfer results.}
\label{tab:s2-fwt}
\end{table}

\begin{figure}[t]
\centering
\setlength{\abovecaptionskip}{0.2cm}
\setlength{\belowcaptionskip}{-0.4cm}
\includegraphics[width=\columnwidth]{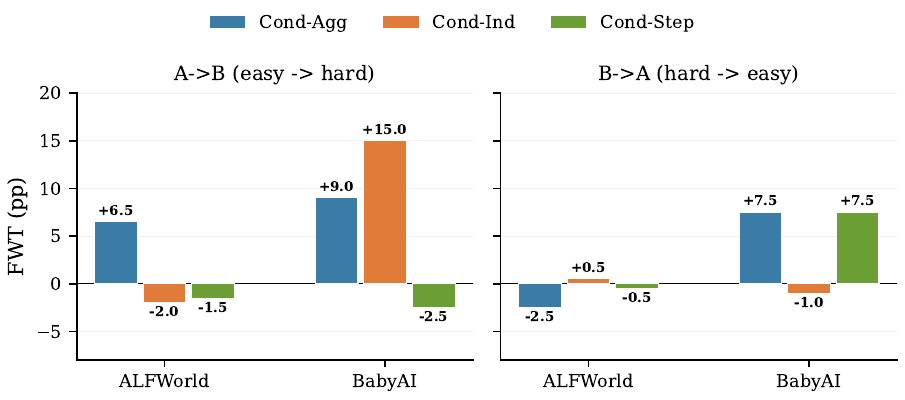}
\caption{FWT by memory organization. No condition universally dominates; the best configuration depends on the task pair.}
\label{fig:s2-fwt}
\end{figure}

\begin{figure}[t]
\centering
\setlength{\abovecaptionskip}{0.2cm}
\setlength{\belowcaptionskip}{-0.4cm}
\includegraphics[width=\columnwidth]{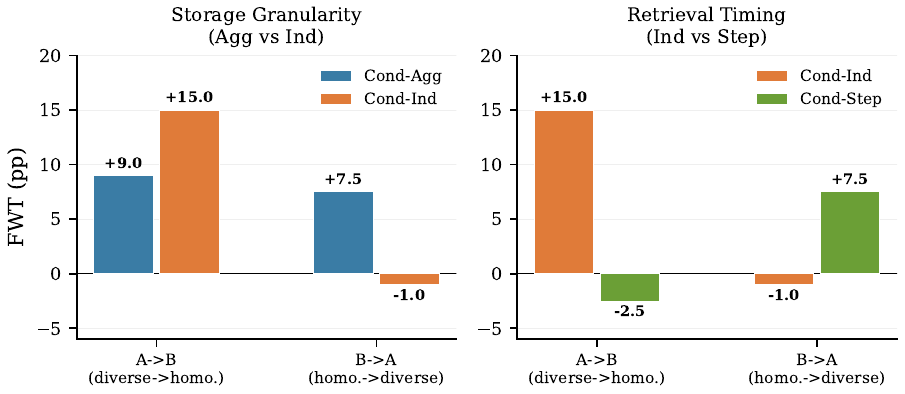}
\caption{BabyAI: FWT by direction, contrasting unit granularity (left) and retrieval frequency (right). Fine-grained storage or retrieval helps in one direction but not the other.}
\label{fig:s2-interaction}
\end{figure}

\subsection{Memory-Unit Granularity (Agg vs.\ Ind)}

Tables~\ref{tab:s2-fwt} and Figure~\ref{fig:s2-fwt} present the results.
The key pattern is that \textbf{finer storage granularity helps only when splitting memory produces entries that are genuinely different in content and can be retrieved separately.}

\paragraph{BabyAI: Diverse And Homogeneous Source Memories Behave Differently.}
The two directions produce opposite results (Figure~\ref{fig:s2-interaction}, left).

\begin{figure}[t]
\centering
\setlength{\abovecaptionskip}{0.2cm}
\setlength{\belowcaptionskip}{-0.4cm}
\includegraphics[width=\columnwidth]{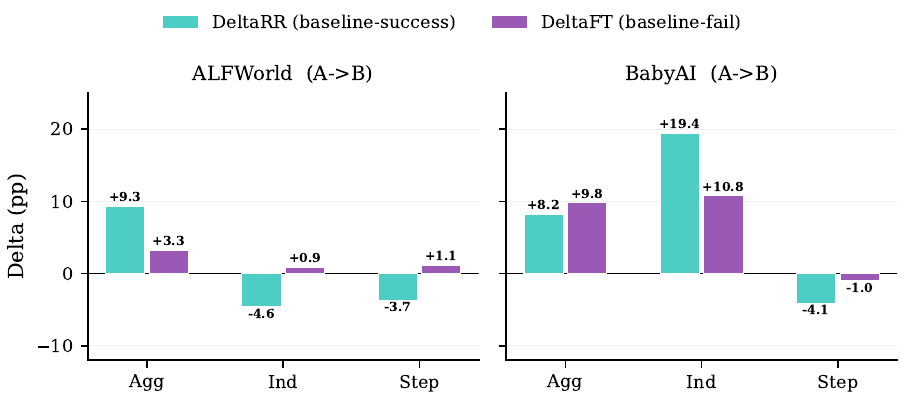}
\caption{$\drr$/$\dft$ diagnostic for A$\to$B (Study~2). On BabyAI, Cond-Ind shows broad improvement across both subsets ($\drr = +19.4$, $\dft = +10.8$), while on ALFWorld Cond-Agg yields positive diagnostics.}
\label{fig:s2-rrft}
\end{figure}

In A$\to$B, Cond-Ind achieves $+$15.0\% (vs.\ Agg's $+$9.0\%), with both subsets improving ($\drr = +19.4$, $\dft = +10.8$; Figure~\ref{fig:s2-rrft}).
Task~A contains three distinct sub-task types (go-to, pick-up, open-door), so its insights reflect different strategies.
When stored individually, each entry has a distinct retrieval key, allowing the retriever to match Task~B queries to the most relevant one.

In B$\to$A, Cond-Ind drops to $-$1.0\% while Agg reaches $+$7.5\%.
Task~B episodes are essentially systematic grid search (“find [object]”), producing highly repetitive insights.
Splitting them creates a large pool of near-identical entries that the retriever cannot meaningfully distinguish.
Individually indexing identical memories therefore offers little advantage over bundling.
We call this \emph{retrieval diversity collapse}: the pool appears large, but homogeneous queries repeatedly retrieve the same few entries (Appendix~\ref{app:diversity-collapse}).
Appendix~\ref{app:case} provides retrieval-level examples of both regimes.

\paragraph{ALFWorld: Bundling Remains The Safer Choice.}
ALFWorld shows a milder contrast than BabyAI, but the pattern is compatible with the same conclusion.
In A$\to$B, Agg outperforms Ind ($+$6.5 vs.\ $-$2.0), and only Agg yields consistently positive diagnostics ($\drr = +9.3$, $\dft = +3.3$; Figure~\ref{fig:s2-rrft}).
The results are consistent with the view that ALFWorld tasks rely on coherent multi-step strategies (e.g., find$\to$take$\to$clean$\to$place), so fragmenting those strategies into individually indexed insights may weaken the structure that is preserved when they are bundled.

\paragraph{Overall.}
The two environments point to the same practical lesson from different angles.
Fine-grained storage is most useful when task experience can be decomposed into self-contained insights that remain useful on their own.
When the stored experience is highly repetitive or is better understood as one coherent multi-step strategy, splitting it into many entries is less helpful, and coarser aggregation is the safer choice.

\subsection{Retrieval Frequency (Ind vs.\ Step)}

The previous section examined how memories should be stored; here we ask how often they should be retrieved.
The key question is whether step-level retrieval helps when an agent's information needs change during an episode.
We use the Ind--Step comparison to study retrieval frequency.
Appendix~\ref{app:case} provides relevant case analyses.

\paragraph{BabyAI: Step Helps When Tasks Have Distinct Phases.}
In B$\to$A, Step achieves $+$7.5\% FWT while Ind drops to $-$1.0\%.
Task~A contains distinct execution phases: the agent must navigate and then interact with objects or doors.
Periodic re-querying may help because the information needed later in the episode differs from what is needed initially.

In A$\to$B, the pattern reverses: Step yields $-$2.5\% while Ind achieves $+$15.0\%.
Task~B follows a largely uniform execution pattern, essentially systematic grid search throughout the episode.
When information needs remain stable over time, re-querying is less likely to surface new guidance and more likely to return redundant content.

\paragraph{ALFWorld: Within-Task Gains Do Not Transfer Cleanly.}
On ALFWorld, cross-task FWT differences between Ind and Step are small ($\pm$1.5pp), making retrieval-frequency effects less distinct under task shift.
However, Step achieves substantially higher \emph{within-task} performance (e.g., 79.5\% vs.\ 55.0\% scratch on Task~B; Appendix~\ref{app:step-analysis}).
This suggests step-level retrieval can be highly effective for multi-step tasks within the same distribution, but the advantage does not necessarily transfer across tasks.
As in Study~1, within-task performance alone may therefore give a misleading picture of which memory design is best under task shift.

\paragraph{Overall.}
More frequent retrieval is not universally beneficial.
It appears most useful when the target task's information needs change over the course of execution; when they remain stable, a single initial retrieval may be enough.

\begin{figure}[t]
\centering
\includegraphics[width=\columnwidth]{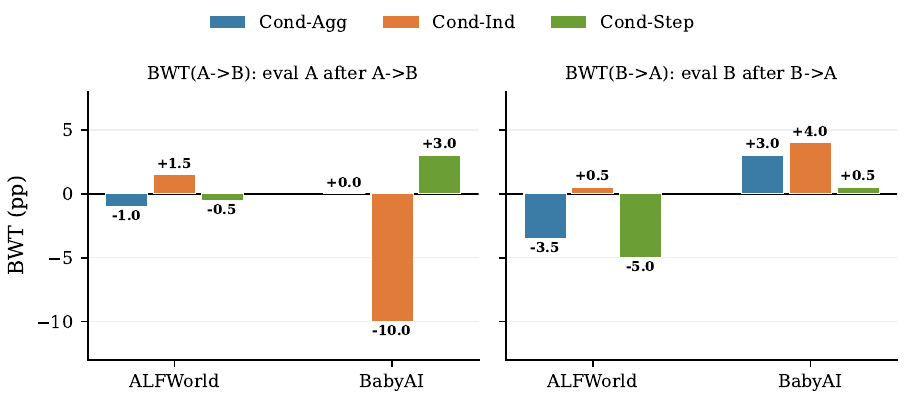}
\caption{BWT by memory organization. On BabyAI under the A$\to$B sequence, Cond-Ind yields $-$10\% (forgetting), while Cond-Step achieves $+$3\% (positive backward transfer).}
\label{fig:s2-bwt}
\end{figure}

\begin{table}[t]
\centering
\small
\setlength{\tabcolsep}{3.5pt}
\begin{tabular}{ll rr rr}
\toprule
& & \multicolumn{2}{c}{\textbf{BWT(A$\to$B)}} & \multicolumn{2}{c}{\textbf{BWT(B$\to$A)}} \\
\cmidrule(lr){3-4} \cmidrule(lr){5-6}
\textbf{Env} & \textbf{Cond} & BWT & $\drr$ & BWT & $\drr$ \\
\midrule
\multirow{3}{*}{ALF}
& Agg  & $-$1.0 & ~~0.0  & $-$3.5 & $+$4.6 \\
& Ind  & $+$1.5 & $+$1.1 & $+$0.5 & $-$0.9 \\
& Step & $-$0.5 & $-$0.5 & $-$5.0 & $-$4.6 \\
\midrule
\multirow{3}{*}{Baby}
& Agg  & ~~0.0  & $+$2.2 & $+$3.0 & $+$8.2 \\
& Ind  & $-$10.0 & $-$6.4 & $+$4.0 & $+$7.1 \\
& Step & $+$3.0 & $-$2.2 & $+$0.5 & $-$5.1 \\
\bottomrule
\end{tabular}
\caption{Study~2 backward transfer result.}
\label{tab:s2-bwt}
\end{table}

\subsection{Stability-Plasticity Trade-off}

While forward transfer suggests that fine-grained storage can be effective, backward transfer reveals an important caveat (Table~\ref{tab:s2-bwt}, Figure~\ref{fig:s2-bwt}).

The clearest case appears in BabyAI under the A$\to$B sequence: Cond-Ind achieves the strongest forward transfer but also the most severe forgetting.
In the same run, the design that adapts best to the new task degrades most on the old one.
This pattern reflects a \textbf{stability-plasticity trade-off}, but the mechanism here is not parameter overwriting but \textbf{memory dilution}.

After Task~B training, the memory pool becomes dominated by near-identical grid-search insights.
When re-evaluated on Task~A, the agent's queries often retrieve these homogeneous new entries instead of the original Task~A memories.
Old memories are not erased but become harder to access as the pool grows (Appendix~\ref{app:case}).

Cond-Agg mitigates this issue by keeping the pool compact through aggregation.
Cond-Step may partially compensate via step-level retrieval, which can surface different items across execution steps even in a diluted pool.
On ALFWorld, BWT values are small across conditions, making this sequence less informative about stability differences.

Overall, these results suggest that external memory may not eliminate the stability-plasticity dilemma but shift its mechanism from parameter updates to retrieval dynamics.

\subsection{Study 2 Summary}

\begin{enumerate}[nosep,leftmargin=*]
\item \textbf{Finer storage granularity is not always better.}
Indexing insights individually helps when source memories are diverse enough for the retriever to discriminate, but hurts when memories are homogeneous or tasks require coherent multi-step guidance.

\item \textbf{More frequent retrieval is not universally better either.}
Periodic re-querying helps when tasks contain distinct execution phases, but adds noise when execution is uniform throughout.

\item \textbf{Retrieval diversity collapse explains why finer granularity can fail.}
Fine-grained storage creates a seemingly rich pool, but homogeneous queries funnel most retrievals to the similar few items (Appendix~\ref{app:diversity-collapse}).

\item \textbf{A stability-plasticity trade-off can emerge through memory dilution.}
The condition with the strong forward transfer also shows the significant forgetting, suggesting that the dilemma can emerge in non-parametric CL through retrieval dynamics rather than parameter overwriting.
\end{enumerate}

\section{Related Work}
\label{sec:related}

\subsection{Agent Experience and Memory}
A growing line of work equips LLM agents with external memory so they can improve through interaction without updating model weights. These systems mainly differ in what they treat as the reusable unit of \emph{experience}: reflections or lessons \citep{shinn2023reflexion,zhao2024expel}, executable skills or workflows \citep{wang2023voyager,nottingham2024sso,wang2024awm,zhang2026memskill}, full trajectories or exemplars \citep{zheng2024synapse}, and persistent abstractions or memory structures \citep{majumder2023clin,park2023generative,packer2023memgpt,zhong2024memorybank}. Recent work also studies cross-task reuse and memory-side failure modes \citep{li2025mael,xiong2025experience}. We examine what these memory designs imply under task shift.

\subsection{Continual Learning for LLMs and Agents}
Classical continual learning studies how to balance stability and plasticity under non-stationary data through regularization, replay, and architectural isolation or expansion \citep{kirkpatrick2017ewc,li2016lwf,lopez2017gradient,mallya2018packnet}, with recent surveys extending this perspective to foundation models \citep{wang2024clsurvey,wu2024llmcl}. For LLMs, continual or instruction tuning remains vulnerable to forgetting, motivating approaches such as non-parametric memory, parameter-efficient adaptation, expert composition, and architectural preservation \citep{luo2023empirical,gutierrez2025ragmemory,wang2025gorp,wang2025see,biswas2026ella}. Continual learning is also studied at the agent level, e.g., through disentangled parameter updates in Agent-Dice \citep{wu2026agentdice}. We focus on the complementary frozen-backbone setting, where interference arises in retrieval rather than parameter updates.
\section{Conclusion}

External memory may appear to sidestep the stability-plasticity dilemma because old experience is preserved while new experience can always be appended.
Our results suggest a more cautious view: once behavior depends on retrieval through a finite context window, old and new memories compete for the same channel into the agent's decisions, reintroducing transfer and interference through retrieval rather than through parameter updates.
Across both studies, abstraction shapes whether retrieved experience helps or misleads, memory granularity determines whether stored diversity becomes useful guidance or repetitive noise, and the resulting system remains subject to trade-offs that require careful design.
\section*{Limitations}

Our experiments prioritize controlled analysis over broad coverage. In practice, even a single task-pair study expands into many training and evaluation groups once baselines, single-task controls, both transfer directions, backward probes, repeated runs, and multiple memory conditions are included. For this reason, we focus on two environments with one representative task pair each, one sparse retrieval backbone, and two runs per condition. We do not claim this setup is exhaustive; rather, we view it as a resource-constrained but targeted testbed that isolates key design dimensions and makes the resulting transfer, forgetting, and retrieval effects easier to interpret. Broader task coverage, additional retrievers, and larger-scale repetition would strengthen the generality of the conclusions and are important directions for future improvements.


\bibliography{custom}

@article{shinn2023reflexion,
  title={Reflexion: Language agents with verbal reinforcement learning},
  author={Shinn, Noah and Cassano, Federico and Gopinath, Ashwin and Narasimhan, Karthik and Yao, Shunyu},
  journal={Advances in neural information processing systems},
  volume={36},
  pages={8634--8652},
  year={2023}
}

@article{wang2023voyager,
  title={Voyager: An open-ended embodied agent with large language models},
  author={Wang, Guanzhi and Xie, Yuqi and Jiang, Yunfan and Mandlekar, Ajay and Xiao, Chaowei and Zhu, Yuke and Fan, Linxi and Anandkumar, Anima},
  journal={arXiv preprint arXiv:2305.16291},
  year={2023}
}

@inproceedings{zhao2024expel,
  title={Expel: Llm agents are experiential learners},
  author={Zhao, Andrew and Huang, Daniel and Xu, Quentin and Lin, Matthieu and Liu, Yong-Jin and Huang, Gao},
  booktitle={Proceedings of the AAAI Conference on Artificial Intelligence},
  volume={38},
  number={17},
  pages={19632--19642},
  year={2024}
}

@inproceedings{zhong2024memorybank,
  title={Memorybank: Enhancing large language models with long-term memory},
  author={Zhong, Wanjun and Guo, Lianghong and Gao, Qiqi and Ye, He and Wang, Yanlin},
  booktitle={Proceedings of the AAAI conference on artificial intelligence},
  volume={38},
  number={17},
  pages={19724--19731},
  year={2024}
}

@article{shridhar2021alfworld,
  title={Alfworld: Aligning text and embodied environments for interactive learning},
  author={Shridhar, Mohit and Yuan, Xingdi and C{\^o}t{\'e}, Marc-Alexandre and Bisk, Yonatan and Trischler, Adam and Hausknecht, Matthew},
  journal={arXiv preprint arXiv:2010.03768},
  year={2020}
}

@article{chevalier2019babyai,
  title={Babyai: A platform to study the sample efficiency of grounded language learning},
  author={Chevalier-Boisvert, Maxime and Bahdanau, Dzmitry and Lahlou, Salem and Willems, Lucas and Saharia, Chitwan and Nguyen, Thien Huu and Bengio, Yoshua},
  journal={arXiv preprint arXiv:1810.08272},
  year={2018}
}

@article{kirkpatrick2017ewc,
  title={Overcoming catastrophic forgetting in neural networks},
  author={Kirkpatrick, James and Pascanu, Razvan and Rabinowitz, Neil and Veness, Joel and Desjardins, Guillaume and Rusu, Andrei A and Milan, Kieran and Quan, John and Ramalho, Tiago and Grabska-Barwinska, Agnieszka and others},
  journal={Proceedings of the national academy of sciences},
  volume={114},
  number={13},
  pages={3521--3526},
  year={2017},
  publisher={National Academy of Sciences}
}

@article{lopez2017gradient,
  title={Gradient episodic memory for continual learning},
  author={Lopez-Paz, David and Ranzato, Marc'Aurelio},
  journal={Advances in neural information processing systems},
  volume={30},
  year={2017}
}

@article{wang2024clsurvey,
  title={A comprehensive survey of continual learning: Theory, method and application},
  author={Wang, Liyuan and Zhang, Xingxing and Su, Hang and Zhu, Jun},
  journal={IEEE transactions on pattern analysis and machine intelligence},
  volume={46},
  number={8},
  pages={5362--5383},
  year={2024},
  publisher={IEEE}
}

@inproceedings{yao2023react,
  title={React: Synergizing reasoning and acting in language models},
  author={Yao, Shunyu and Zhao, Jeffrey and Yu, Dian and Du, Nan and Shafran, Izhak and Narasimhan, Karthik R and Cao, Yuan},
  booktitle={The eleventh international conference on learning representations},
  year={2022}
}

@book{robertson2009bm25,
  title={The probabilistic relevance framework: BM25 and beyond},
  author={Robertson, Stephen and Zaragoza, Hugo},
  volume={4},
  year={2009},
  publisher={Now Publishers Inc}
}

@article{xiong2024agentgym,
      title={AgentGym: Evolving Large Language Model-based Agents across Diverse Environments}, 
      author={Zhiheng Xi and Yiwen Ding and Wenxiang Chen and Boyang Hong and Honglin Guo and Junzhe Wang and Dingwen Yang and Chenyang Liao and Xin Guo and Wei He and Songyang Gao and Lu Chen and Rui Zheng and Yicheng Zou and Tao Gui and Qi Zhang and Xipeng Qiu and Xuanjing Huang and Zuxuan Wu and Yu-Gang Jiang},
      journal={arXiv preprint arXiv:2406.04151},
      year={2024},
}

@article{packer2023memgpt,
  title={MemGPT: towards LLMs as operating systems.},
  author={Packer, Charles and Fang, Vivian and Patil, Shishir\_G and Lin, Kevin and Wooders, Sarah and Gonzalez, Joseph\_E},
  year={2023},
  journal={ArXiv}
}

@inproceedings{park2023generative,
  title={Generative agents: Interactive simulacra of human behavior},
  author={Park, Joon Sung and O'Brien, Joseph and Cai, Carrie Jun and Morris, Meredith Ringel and Liang, Percy and Bernstein, Michael S},
  booktitle={Proceedings of the 36th annual acm symposium on user interface software and technology},
  pages={1--22},
  year={2023}
}

@article{wang2023longmem,
  title={Augmenting language models with long-term memory},
  author={Wang, Weizhi and Dong, Li and Cheng, Hao and Liu, Xiaodong and Yan, Xifeng and Gao, Jianfeng and Wei, Furu},
  journal={Advances in Neural Information Processing Systems},
  volume={36},
  pages={74530--74543},
  year={2023}
}

@article{luo2023empirical,
  title={An empirical study of catastrophic forgetting in large language models during continual fine-tuning},
  author={Luo, Yun and Yang, Zhen and Meng, Fandong and Li, Yafu and Zhou, Jie and Zhang, Yue},
  journal={IEEE Transactions on Audio, Speech and Language Processing},
  year={2025},
  publisher={IEEE}
}

@article{xu2025amem,
  title={A-mem: Agentic memory for llm agents},
  author={Xu, Wujiang and Liang, Zujie and Mei, Kai and Gao, Hang and Tan, Juntao and Zhang, Yongfeng},
  journal={arXiv preprint arXiv:2502.12110},
  year={2025}
}

@article{xiong2025experience,
  title={How memory management impacts llm agents: An empirical study of experience-following behavior},
  author={Xiong, Zidi and Lin, Yuping and Xie, Wenya and He, Pengfei and Liu, Zirui and Tang, Jiliang and Lakkaraju, Himabindu and Xiang, Zhen},
  journal={arXiv preprint arXiv:2505.16067},
  year={2025}
}

@article{gutierrez2025ragmemory,
  title={From rag to memory: Non-parametric continual learning for large language models},
  author={Guti{\'e}rrez, Bernal Jim{\'e}nez and Shu, Yiheng and Qi, Weijian and Zhou, Sizhe and Su, Yu},
  journal={arXiv preprint arXiv:2502.14802},
  year={2025}
}

@article{wu2026agentdice,
  title={Agent-Dice: Disentangling Knowledge Updates via Geometric Consensus for Agent Continual Learning},
  author={Wu, Zheng and Lou, Xingyu and Ma, Xinbei and Li, Yansi and Liu, Weiwen and Zhang, Weinan and Wang, Jun and Zhang, Zhuosheng},
  journal={arXiv preprint arXiv:2601.03641},
  year={2026}
}

@article{wu2024llmcl,
  title={Continual learning for large language models: A survey},
  author={Wu, Tongtong and Luo, Linhao and Li, Yuan-Fang and Pan, Shirui and Vu, Thuy-Trang and Haffari, Gholamreza},
  journal={arXiv preprint arXiv:2402.01364},
  year={2024}
}

@article{cao2025remember,
  title={Remember me, refine me: A dynamic procedural memory framework for experience-driven agent evolution},
  author={Cao, Zouying and Deng, Jiaji and Yu, Li and Zhou, Weikang and Liu, Zhaoyang and Ding, Bolin and Zhao, Hai},
  journal={arXiv preprint arXiv:2512.10696},
  year={2025}
}

@inproceedings{borgeaud2022retro,
  title={Improving language models by retrieving from trillions of tokens},
  author={Borgeaud, Sebastian and Mensch, Arthur and Hoffmann, Jordan and Cai, Trevor and Rutherford, Eliza and Millican, Katie and Van Den Driessche, George Bm and Lespiau, Jean-Baptiste and Damoc, Bogdan and Clark, Aidan and others},
  booktitle={International conference on machine learning},
  pages={2206--2240},
  year={2022},
  organization={PMLR}
}

@article{zhang2025memorysurvey,
  title={A survey on the memory mechanism of large language model-based agents},
  author={Zhang, Zeyu and Dai, Quanyu and Bo, Xiaohe and Ma, Chen and Li, Rui and Chen, Xu and Zhu, Jieming and Dong, Zhenhua and Wen, Ji-Rong},
  journal={ACM Transactions on Information Systems},
  volume={43},
  number={6},
  pages={1--47},
  year={2025},
  publisher={ACM New York, NY}
}

@article{majumder2023clin,
  title={CLIN: A Continually Learning Language Agent for Rapid Task Adaptation and Generalization},
  author={Majumder, Bodhisattwa Prasad and Dalvi Mishra, Bhavana and Jansen, Peter and Tafjord, Oyvind and Tandon, Niket and Zhang, Li and Callison-Burch, Chris and Clark, Peter},
  journal={arXiv preprint arXiv:2310.10134},
  year={2023}
}

@article{zheng2024synapse,
  title={Synapse: Trajectory-as-exemplar prompting with memory for computer control},
  author={Zheng, Longtao and Wang, Rundong and Wang, Xinrun and An, Bo},
  journal={arXiv preprint arXiv:2306.07863},
  year={2023}
}

@article{nottingham2024sso,
  title={Skill set optimization: Reinforcing language model behavior via transferable skills},
  author={Nottingham, Kolby and Majumder, Bodhisattwa Prasad and Mishra, Bhavana Dalvi and Singh, Sameer and Clark, Peter and Fox, Roy},
  journal={arXiv preprint arXiv:2402.03244},
  year={2024}
}

@article{wang2024awm,
  title={Agent workflow memory},
  author={Wang, Zora Zhiruo and Mao, Jiayuan and Fried, Daniel and Neubig, Graham},
  journal={arXiv preprint arXiv:2409.07429},
  year={2024}
}

@article{li2025mael,
  title={Cross-Task Experiential Learning on LLM-based Multi-Agent Collaboration},
  author={Li, Yilong and Qian, Chen and Xia, Yu and Shi, Ruijie and Dang, Yufan and Xie, Zihao and You, Ziming and Chen, Weize and Yang, Cheng and Liu, Weichuan and others},
  journal={arXiv preprint arXiv:2505.23187},
  year={2025}
}

@article{li2016lwf,
  title={Learning without forgetting},
  author={Li, Zhizhong and Hoiem, Derek},
  journal={IEEE transactions on pattern analysis and machine intelligence},
  volume={40},
  number={12},
  pages={2935--2947},
  year={2017},
  publisher={IEEE}
}

@inproceedings{mallya2018packnet,
  title={Packnet: Adding multiple tasks to a single network by iterative pruning},
  author={Mallya, Arun and Lazebnik, Svetlana},
  booktitle={Proceedings of the IEEE conference on Computer Vision and Pattern Recognition},
  pages={7765--7773},
  year={2018}
}

@inproceedings{wang2025gorp,
  title={Continual gradient low-rank projection fine-tuning for LLMs},
  author={Wang, Chenxu and Lyu, Yilin and Sun, Zicheng and Jing, Liping},
  booktitle={Proceedings of the 63rd Annual Meeting of the Association for Computational Linguistics (Volume 1: Long Papers)},
  pages={14815--14829},
  year={2025}
}

@inproceedings{wang2025see,
  title={SEE: Continual Fine-tuning with Sequential Ensemble of Experts},
  author={Wang, Zhilin and Li, Yafu and Qu, Xiaoye and Cheng, Yu},
  booktitle={Findings of the Association for Computational Linguistics: ACL 2025},
  pages={7418--7432},
  year={2025}
}

@article{biswas2026ella,
  title={ELLA: Efficient lifelong learning for adapters in large language models},
  author={Biswas, Shristi Das and Zhang, Yue and Pal, Anwesan and Bhargava, Radhika and Roy, Kaushik},
  journal={arXiv preprint arXiv:2601.02232},
  year={2026}
}

@article{zhang2026memskill,
  title={MemSkill: Learning and Evolving Memory Skills for Self-Evolving Agents},
  author={Zhang, Haozhen and Long, Quanyu and Bao, Jianzhu and Feng, Tao and Zhang, Weizhi and Yue, Haodong and Wang, Wenya},
  journal={arXiv preprint arXiv:2602.02474},
  year={2026}
}

\appendix

\section{Experimental Details}
\label{app:details}

\subsection{Terminology and Conditions}

\begin{table}[h]
\centering
\small
\setlength{\tabcolsep}{3pt}
\begin{tabular}{lp{0.68\columnwidth}}
\toprule
\textbf{Term} & \textbf{Meaning in this paper} \\
\midrule
Raw & Full action--observation trajectory from one episode. \\
Insight & LLM-distilled strategy insights derived from a trajectory. \\
Cond-Agg & Task-level aggregation of insights, retrieved once per episode. \\
Cond-Ind & Insight-level storage, with each insight indexed separately and retrieved once per episode. \\
Cond-Step & Same insight-level storage as Cond-Ind, but with step-level retrieval; in our implementation, the agent re-queries periodically during execution. \\
\bottomrule
\end{tabular}
\caption{Representation and condition glossary.}
\label{tab:appendix-glossary}
\end{table}

\subsubsection{Run Types and Labels}
\textbf{No-memory baseline} disables external memory entirely.
\textbf{Scratch run} means a memory-enabled agent trained on a single task family from empty memory.
\textbf{Cross-task run} means a memory-enabled agent trained on one task family and then continued on another while reusing the accumulated pool.
\textbf{Eval-only probe} denotes post-training evaluation on the earlier task for backward transfer.
Throughout the paper, \textbf{FWT($X\to Y$)} refers to forward transfer on the later task $Y$, while \textbf{BWT($X\to Y$)} refers to backward transfer measured by re-evaluating the earlier task $X$ after the full $X\to Y$ sequence.

\subsubsection{Metrics and Diagnostics}
\textbf{FWT($X\to Y$)} is defined as $\text{Acc}_Y(\varnothing\!\to\!X\!\to\!Y)-\text{Acc}_Y(\varnothing\!\to\!Y)$ and measures whether prior experience from $X$ helps or hurts learning on the later task $Y$.
\textbf{BWT($X\to Y$)} is defined as $\text{Acc}_X(\varnothing\!\to\!X\!\to\!Y)-\text{Acc}_X(\varnothing\!\to\!X)$ and measures whether the later phase changes performance on the earlier task $X$.
\textbf{RR} (Retention Rate) is the success rate on the baseline-success subset after memory is introduced, and \textbf{NL} (New Learning rate) is the success rate on the baseline-fail subset after memory is introduced.
For a cross-task run $X\to T$, we report $\drr = \text{RR}(\varnothing\!\to\!X\!\to\!T)-\text{RR}(\varnothing\!\to\!T)$ and $\dft = \text{NL}(\varnothing\!\to\!X\!\to\!T)-\text{NL}(\varnothing\!\to\!T)$.

\subsubsection{Memory System}
We build on AgentGym \citep{xiong2024agentgym} with the ReMe memory module \citep{cao2025remember}.
Retrieval uses BM25 \citep{robertson2009bm25} throughout.

\subsubsection{Study~1 Conditions}
Both representations use instance-level storage with a single retrieved memory per query (top-1).
\textbf{Raw} stores the complete action--observation trajectory from each episode.
\textbf{Insight} prompts the LLM to distill the trajectory into abstract strategy insights (${\sim}$3 per instance), which are aggregated into a single task-level memory unit keyed by the original task instruction.

\subsubsection{Study~2 Conditions}
All three conditions use the Insight representation.
\textbf{Cond-Agg} stores one task-level bundle per instance (key = task instruction) and retrieves the top-1 match at the start of each episode (${\sim}$3 insights per retrieval).
\textbf{Cond-Ind} stores each insight as a separate memory unit (key = a \texttt{when\_to\_use} descriptor generated alongside the insight) and retrieves the top-3 matches at the start of each episode.
\textbf{Cond-Step} uses the same insight-level storage as Cond-Ind but re-queries every 4 execution steps, allowing dynamically updated guidance.
All three conditions thus surface approximately 3 insights per retrieval event, holding the information budget roughly constant.

\subsubsection{Training and Evaluation}
All conditions use Qwen-Plus as the backbone LLM.
Each task phase contains 200 training instances and 100 test instances.
Results are averaged over two independent runs.

\subsection{Hyperparameters}
\label{app:hyperparams}

Table~\ref{tab:appendix-hyperparams} summarizes the main hyperparameters used in the experiments.

\begin{table*}[t]
\centering
\small
\setlength{\tabcolsep}{4pt}
\begin{tabular}{lp{0.22\textwidth}p{0.14\textwidth}p{0.40\textwidth}}
\toprule
\textbf{Scope} & \textbf{Hyperparameter} & \textbf{Value} & \textbf{Notes} \\
\midrule
Shared & Backbone LLM & Qwen-Plus & Used for both training and evaluation across all reported conditions. \\
Shared & Training instances per task phase & 200 & Matches \texttt{train\_size} in the experiment runner. \\
Shared & Test instances per task phase & 100 & Fixed task split used in both environments. \\
Shared & Number of runs & 2 & Final results are averaged over two independent runs. \\
Shared & Random seed & 42 & Set in the launch scripts. \\
Shared & Evaluation workers & 4 & \texttt{EVAL\_PROCESSES=4} in the chapter scripts. \\
Shared & Milestones & 1 & One end-of-phase evaluation checkpoint per task phase. \\
Shared & Max generation tokens & 4096 & Overrides the lower parser default in the launcher configuration. \\
Shared & Temperature & 0.5 & Shared sampling setting in the chapter scripts. \\
Study~1 & Memory representation & Raw / Insight & Corresponds to trajectory memory vs.\ distilled insight memory. \\
Study~1 & Retrieval top-$k$ & 1 & Single retrieved memory per query. \\
Study~1 & Success-only filtering & false & Both successful and unsuccessful episodes may enter memory. \\
Study~2 & Conditions & Cond-Agg / Cond-Ind / Cond-Step & All three use the Insight representation but differ in organization and retrieval timing. \\
Study~2 & Retrieval top-$k$ & 1 / 3 / 3 & Cond-Agg retrieves one aggregated bundle; Cond-Ind and Cond-Step retrieve top-3 insight entries. \\
Study~2 & Step retrieval interval & 4 steps & Cond-Step re-queries memory every 4 environment steps. \\
Study~2 & Success-only filtering & false & Same memory inclusion rule as in Study~1. \\
\bottomrule
\end{tabular}
\caption{Main hyperparameters used in our experiments.}
\label{tab:appendix-hyperparams}
\end{table*}

\section{Memory Content Examples}
\label{app:examples}

We show representative entries for each memory type studied in this paper.
All examples are drawn from actual experimental runs and lightly abbreviated for space.

\subsection{Study~1: Raw vs.\ Insight}

\noindent\textbf{Raw entry} (ALFWorld, ``put some bowl on fridge'').\\
\emph{Key}: the full task instruction plus initial environment observation.\\
\emph{Content}: the complete action--observation trajectory from the episode, including all agent reasoning steps and environment responses.
For example:
\begin{quote}\small
\texttt{[User]:} Your task is: put some bowl on fridge.\\
\texttt{[Agent]:} I need to find a bowl first. Let me check the countertop\ldots\\
\texttt{Action:} go to countertop 1\\
\texttt{[User]:} On the countertop 1, you see a bowl 2\ldots\\
\ldots (\emph{full episode, typically 20--50 turns})
\end{quote}

\noindent\textbf{Insight entry} (ALFWorld, same task).\\
\emph{Key}: same task instruction as Raw.\\
\emph{Content}: $\sim$3 abstract insights distilled by the LLM.
For example:
\begin{quote}\small
\textbf{Insight 1.} Check countertops and tables first, as they are the most common locations for portable objects like bowls and cups.\\
\textbf{Insight 2.} If the target object is not visible, systematically visit receptacles in order of likelihood rather than searching randomly.\\
\textbf{Insight 3.} Once the object is picked up, navigate directly to the target receptacle to avoid unnecessary steps.
\end{quote}

\subsection{Study~2: Cond-Agg vs.\ Cond-Ind}

\noindent\textbf{Cond-Agg entry} (BabyAI, Task~A ``open the purple door'').\\
\emph{Key}: task instruction plus initial observation.\\
\emph{Content}: all $\sim$3 insights bundled together, covering different aspects of the episode (e.g., door traversal strategy, backtracking from dead ends, color-based heuristics).
The entire bundle is retrieved as a single unit.

\vspace{4pt}
\noindent\textbf{Cond-Ind entries} (BabyAI, Task~A, same source episode).\\
Each insight is stored separately with its own retrieval key:
\begin{quote}\small
\textbf{Entry 1.}\\
\emph{Key}: ``When the target object is absent from all observations after multiple reorientations, and multiple closed doors are present''\\
\emph{Content}: ``Prioritize attempting all available closed-door traversals from the initial room before committing to any one path\ldots''

\vspace{2pt}
\textbf{Entry 2.}\\
\emph{Key}: ``After entering an empty room with no objects or doors visible''\\
\emph{Content}: ``An empty room with zero object observations is a strong signal of a dead-end branch; backtracking should be initiated immediately\ldots''
\end{quote}
The retriever can now match an incoming query to the single most relevant entry rather than returning the entire bundle.

\begin{figure*}[ht!]
\centering
\includegraphics[width=0.95\textwidth]{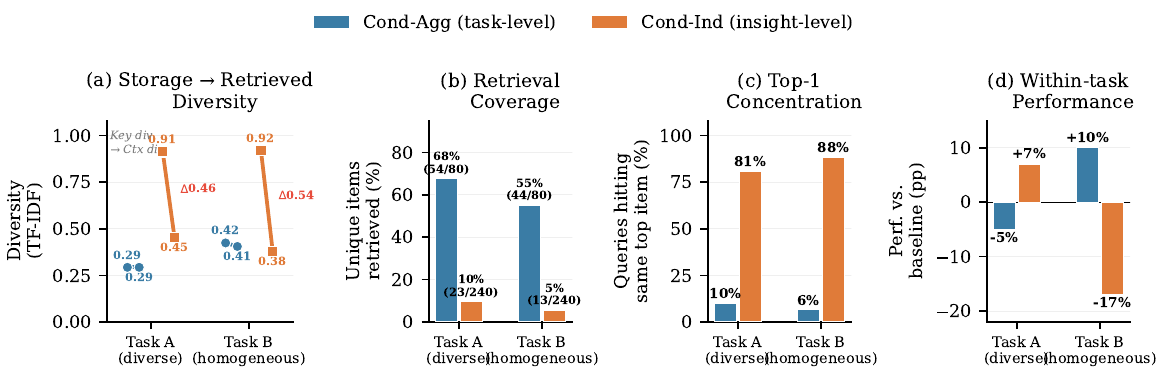}
\caption{Retrieval diversity analysis (BabyAI, within-task setting). \textbf{(a)}~Key diversity in storage (left marker of each pair) vs.\ effective context diversity at retrieval (right marker). For Task~B under Cond-Ind, storage diversity is high ($>$0.91) but context diversity drops to 0.38. \textbf{(b)}~Retrieval coverage: fraction of unique items among all retrievals. Cond-Ind $\times$ Task~B retrieves only 13 unique items out of 240 (5.4\%). \textbf{(c)}~Top-1 concentration: 88.5\% of Task~B queries under Cond-Ind retrieve the identical item. \textbf{(d)}~Within-task performance impact.}
\label{fig:s2-diversity}
\end{figure*}

\section{Retrieval Frequency: Additional Analysis}
\label{app:step-analysis}

\noindent\textbf{Pool content differs between conditions.}\;
Although Cond-Ind and Cond-Step both use insight-level storage, their memory pools are not identical.
During training, the agent retrieves memory before (and, for Step, during) each episode.
Different retrieval affects the agent's behavior, which affects task outcomes, which in turn affects what insights are generated and stored.
Thus the Ind--Step comparison conflates two effects: retrieval frequency at evaluation time and pool composition from training time.

\noindent\textbf{Within-task performance.}\;
Cond-Step achieves the highest within-task (scratch) success rates on ALFWorld: 99.0\% on Task~A and 79.5\% on Task~B, compared to 96.5\%/55.0\% for Cond-Ind and 97.0\%/65.5\% for Cond-Agg.
The advantage is especially large on Task~B ($+$24.5pp over Cond-Ind), indicating that step-level retrieval is highly effective for multi-step manipulation tasks when source and target share the same task distribution.
On BabyAI, Step is middle-ground (70.0\% on A, 48.0\% on B), falling between Ind and Agg.

\noindent\textbf{Scratch baselines affect FWT interpretation.}\;
Because FWT is defined relative to each condition's own scratch performance, different baselines can affect the comparison.
For example, on BabyAI B$\to$A: Cond-Ind's scratch is 75.5\% and cross-task is 74.5\% (FWT $= -1.0$); Cond-Step's scratch is 70.0\% and cross-task is 77.5\% (FWT $= +7.5$).
Step achieves higher \emph{absolute} cross-task performance (77.5 vs.\ 74.5), and the FWT gap is amplified by Step's lower scratch baseline.
This does not invalidate the comparison (FWT measures each design's relative gain from cross-task memory), but readers should note that the absolute performance levels are closer than the FWT values suggest.

\section{Retrieval Diversity Collapse}
\label{app:diversity-collapse}

When Cond-Ind is applied to homogeneous source memories (e.g., BabyAI Task~B), individually indexing each insight creates a large pool of near-identical entries.
We call the resulting phenomenon \emph{retrieval diversity collapse}: the pool \emph{appears} diverse from the storage side, but the actual retrieved content is highly repetitive because homogeneous queries cannot discriminate among similar keys.

To quantify this, we measure how diversity changes from storage to retrieval using TF-IDF similarity analysis (Figure~\ref{fig:s2-diversity}).
We compare two levels: \emph{key diversity} (how varied the retrieval keys are in the pool) and \emph{context diversity} (how varied the items actually retrieved across all test queries are).

Under Cond-Agg, key diversity and context diversity are similar (the pool has one entry per task instance, and different queries tend to retrieve different entries).
Under Cond-Ind, the pool contains many individually indexed insights, producing high key diversity ($>$0.91 for both tasks).
However, whether this high key diversity translates into diverse \emph{retrieved} content depends on the queries.

\noindent\textbf{Task~A (heterogeneous queries).}\;
Task~A encompasses three sub-task types, so different test queries activate different subsets of the pool.
Context diversity under Cond-Ind (0.45) exceeds that under Cond-Agg (0.29): individual indexing is helpful here because the retriever can select the most relevant entry per query.

\noindent\textbf{Task~B (homogeneous queries).}\;
Task~B uniformly instructs ``find [object],'' so all test queries are lexically similar.
Despite the pool containing many entries, these similar queries all produce nearly identical retrieval scores, funneling 88.5\% of queries to the same top-1 item (panel~c).
Context diversity under Cond-Ind (0.38) is actually \emph{lower} than under Cond-Agg (0.41), and retrieval coverage drops to 5.4\% (panel~b).
The result is a $-$17pp within-task performance loss (panel~d).

The collapse is invisible from the storage side (key diversity remains high) and manifests only at retrieval time.
This is why pool-side metrics alone (e.g., number of stored entries, key diversity) are insufficient to predict retrieval effectiveness.

\section{Learning Dynamics}
\label{app:dynamics}

\begin{figure*}[h]
\centering
\includegraphics[width=0.95\textwidth]{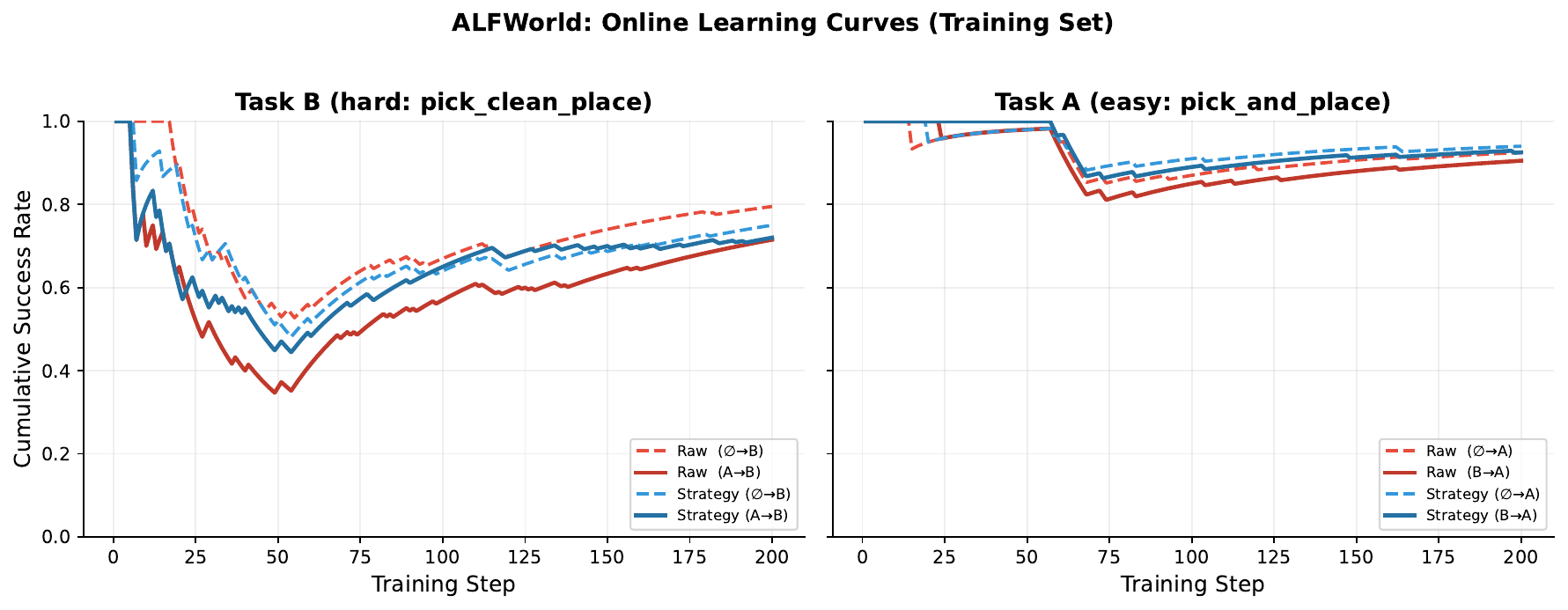}
\caption{ALFWorld cumulative success rate (training set). Raw A$\to$B (solid red) causes a persistent plasticity deficit that never recovers.}
\label{fig:dynamics_csr}
\end{figure*}

Figure~\ref{fig:dynamics_csr} visualizes cumulative success during ALFWorld training for the main Study~1 conditions.
The most important pattern is that the Raw A$\to$B curve drops below the scratch Task~B curve soon after the phase switch and never recovers, indicating that the negative forward-transfer effect is not a brief adaptation cost but a persistent deficit throughout Task~B learning.
By contrast, the Insight A$\to$B curve tracks much closer to, and in some regions above, the scratch baseline, consistent with the interpretation that abstraction makes cross-task reuse safer over the full learning trajectory rather than only at final evaluation.

\begin{figure*}[h]
\centering
\includegraphics[width=0.95\textwidth]{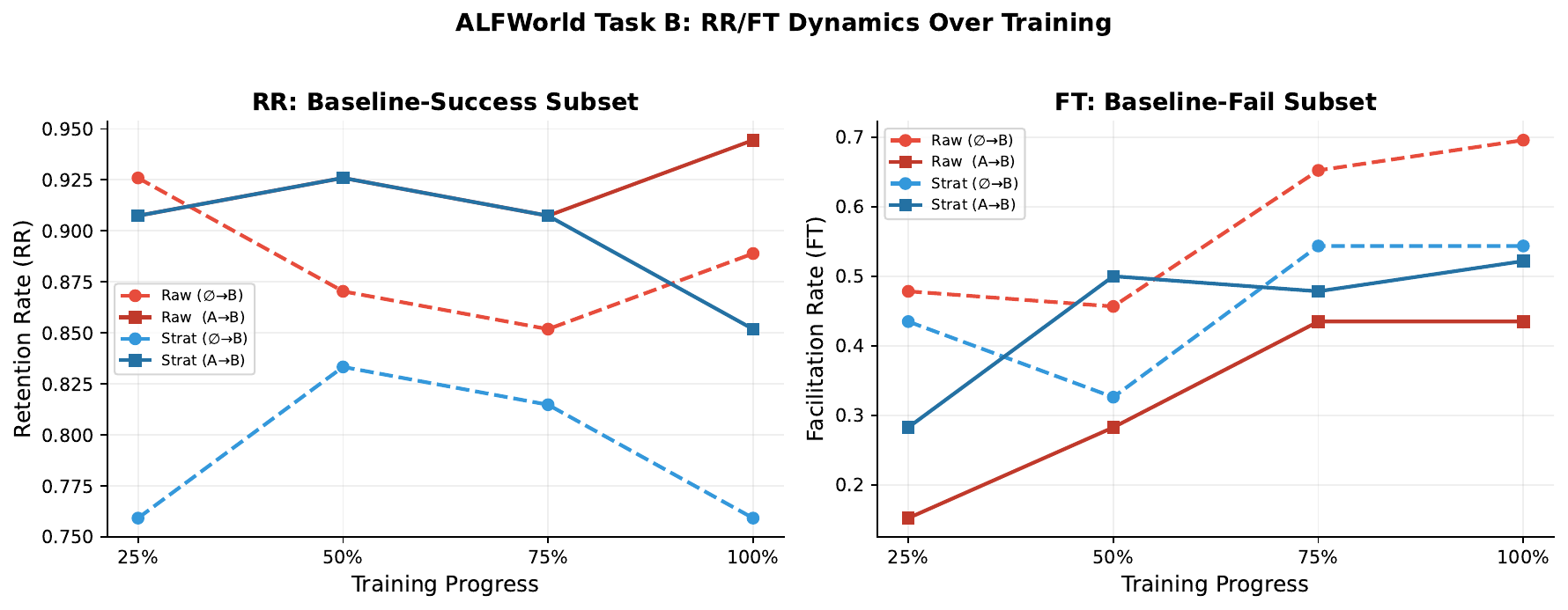}
\caption{ALFWorld Task~B: RR/NL dynamics over training. RR remains stable; the NL gap is the primary driver of negative FWT and persists throughout.}
\label{fig:dynamics_rrft}
\end{figure*}

\section{Case Analyses}
\label{app:case}

\subsection{Study~1: Retrieval Pollution from Raw Trajectories}

 In Study 1, we find that raw trajectories can hurt cross-task reuse because they provide detailed but procedurally wrong guidance, whereas abstract insights are less likely to displace the agent's own reasoning.
The two ALFWorld cases below make that mechanism concrete.

\begin{casebox}{Case 1. Missing cleaning step (ALFWorld)}
\textbf{Task.} ``put a clean bowl in fridge.''

\textbf{Scratch / Insight behavior.}
Both the scratch agent and the insight-based cross-task agent clean the bowl first, then move it to the fridge, and succeed.

\textbf{Raw cross-task retrieval.}
The retrieved memory is a Task~A bowl trajectory whose successful procedure does not require cleaning.

\textbf{Observed reasoning.}
After locating the bowl, the Raw agent states: \emph{``Since bowl~1 was on the diningtable (a clean surface), it's reasonable to assume it's clean.''}
It then moves the bowl directly to the fridge without washing it and fails.

\textbf{Why it matters.}
The retrieved trajectory is locally plausible because it concerns the same object and a very similar placement task, but it omits the one step that distinguishes Task~B from Task~A.
This is exactly the failure mode described in Study~1: raw procedural detail displaces task-specific reasoning under task shift.
\end{casebox}

\begin{casebox}{Case 2. Procedural mimicry (ALFWorld)}
\textbf{Task.} ``clean some cloth and put it in bathtubbasin.''

\textbf{Raw cross-task behavior.}
The Raw agent finds a handtowel, goes to a sinkbasin, searches for soap, places the towel in the sink, and picks it back up, but never issues the correct \texttt{clean handtowel with sinkbasin} command.

\textbf{Insight cross-task retrieval.}
The retrieved guidance is functional rather than procedural: \emph{``only after arriving at the cleaning station did the `clean [object] with [station]' action succeed.''}

\textbf{Outcome.}
The Insight agent issues \texttt{clean cloth with sinkbasin} directly and succeeds.

\textbf{Why it matters.}
This case shows a stronger version of the same phenomenon.
Raw memory does not merely omit a required step; it encourages a longer but functionally wrong action pattern.
Insight memory, by contrast, preserves the condition under which the right action becomes available.
\end{casebox}

Together, these cases support the main Study~1 interpretation: raw trajectories inject task-specific procedures that can \emph{displace} the agent's own reasoning, while abstract insights preserve functional guidance without prescribing a full action script.

\subsection{Study~1: Why the B$\to$A Gap Is Small}
\label{app:s1-direction}

The main text emphasizes the A$\to$B direction because that is where Raw and Insight diverge most sharply.
In the reverse B$\to$A sequence, the gap becomes much smaller.
On BabyAI, both representations yield similar positive forward transfer; on ALFWorld, both are close to zero, in part because Task~A already has a high baseline.

One plausible interpretation is that experience from the relatively harder task is less likely to be actively misleading when reused for the relatively easier one.
In ALFWorld, for example, a ``clean-then-place'' trajectory can still solve the simpler ``place'' task, even if one step is unnecessary.
In BabyAI, Task~B's search-heavy experience can still provide useful navigation guidance for Task~A, even when it does not cover all of Task~A's interaction-specific steps.
Thus the reverse direction appears to involve less harmful procedural mismatch, leaving less for abstraction to correct.

\subsection{Study~2: Granularity and Retrieval-Side Failure Modes}

We conclude three findings from Study~2: fine-grained storage helps when the stored units are genuinely distinguishable, step-level retrieval can sometimes help by aligning guidance with the agent's current execution phase, and backward-transfer failures can arise through memory dilution.
The following cases are designed to illustrate each of these points.

\begin{casebox}{Case 3. Fine-grained storage helps when source memories are genuinely diverse}
\textbf{Setting.}
This case supports `4.1 Memory-Unit Granularity (Agg vs.\ Ind)` on BabyAI Task~A, where Cond-Ind performs well.

\textbf{Representative stored entries.}
In `Task A / Cond-Ind`, the individually indexed insights are qualitatively different.
For example, one key concerns door-adjacent navigation (\emph{``When navigating to a door using `go to <door>' ...''}), another emphasizes preserving the initial spatial anchor (\emph{``Always treat the initial observation as the authoritative spatial anchor ...''}), and another concerns systematic search after repeated failure (\emph{``Persistent absence of a required item ... suggests it is in an unobserved direction ...''}).

\textbf{Representative queries.}
Queries such as \emph{``open the purple door''} and \emph{``go to the red door''} retrieve different top-ranked keys, even though both come from the same task family.

\textbf{Why it matters.}
This is the qualitative pattern behind Cond-Ind's benefit on BabyAI A$\to$B.
Splitting memory into individual entries is useful here because the resulting entries are not merely numerous; they differ in content, and the retriever can select among them.
\end{casebox}

\begin{casebox}{Case 4. Fine-grained storage fails when many entries say roughly the same thing}
\textbf{Setting.}
This case supports `4.1` and complements the retrieval-diversity analysis in Appendix~\ref{app:diversity-collapse}.

\textbf{Representative stored entries.}
In `Task B / Cond-Ind`, many high-scoring keys are variants of the same broad strategy: door traversal, room scanning, and continued search when the target object is absent.
Examples include keys beginning with \emph{``When the initial observation shows only a closed door directly in front ...''} and \emph{``When `go to <obj>' ... loops produce identical empty rooms ...''}.

\textbf{Representative queries.}
Different queries such as \emph{``pick up the ball''} and \emph{``pick up the box''} repeatedly retrieve the same top-ranked key.
In the full retrieval log, the most common top-1 entry is returned for 82 of 200 queries.

\textbf{Why it matters.}
This is the intuitive core of retrieval diversity collapse.
The pool looks large because it contains many individual entries, but many of those entries encode only slight variants of the same search strategy.
In this situation, splitting memory creates redundancy rather than useful granularity.
\end{casebox}

\begin{casebox}{Case 5. Step-level retrieval can align guidance with the current phase}
\textbf{Caveat.}
This case should be read as an illustration of how step-level retrieval can help in practice, not as a perfectly controlled comparison, because Cond-Step and Cond-Ind do not accumulate identical memory pools during training.

\textbf{Representative Step case.}
In one ALFWorld Step run from our main setup, the task is \emph{``clean some egg and put it in sidetable.''}
The retrieved guidance emphasizes the cleaning phase specifically: after acquiring a cleanable object, the agent should first navigate to a sinkbasin because cleaning actions are only available when the agent is co-located with the cleaning fixture.

\textbf{Observed behavior.}
The agent follows this phase-appropriate guidance: it goes to the sinkbasin, executes \texttt{clean egg 1 with sinkbasin 1}, then moves to the sidetable and places the cleaned egg there.

\textbf{Why it matters.}
This case illustrates the kind of guidance a step-style condition can exploit.
The value of more frequent retrieval is not that it is always better, but that it can align guidance with the agent's current execution phase when the task's information needs change over time.
\end{casebox}

\begin{casebox}{Case 6. Memory dilution after homogeneous Task~B training}
\textbf{Setting.}
This case supports `4.3 Backward Transfer and the Stability-Plasticity Trade-off`.

\textbf{Representative pool composition.}
After Task~B training under `Cond-Ind`, the pool is filled with entries that all encode closely related search behavior, for example:
\begin{itemize}[nosep,leftmargin=*]
\item \emph{``When the goal object is absent from all initial observations across multiple rooms ... pivot from blind door progression to systematic orientation-based scanning ...''}
\item \emph{``Persistent linear door traversal without environmental anchoring leads to unbounded exploration loops ...''}
\end{itemize}

\textbf{Why it matters.}
These entries are not wrong for Task~B.
The problem is that, after enough such entries accumulate, they dominate the pool.
Under backward evaluation on Task~A, this makes it harder for more diverse Task~A memories to be retrieved.
The resulting failure is therefore not deletion of old memories, but reduced access to them through a pool dominated by repetitive new guidance.
\end{casebox}

\end{document}